\documentclass[journal,twocolumn,final]{IEEEtran}

\usepackage{cite}
\usepackage[latin9]{inputenc}
\usepackage{amsthm}
\usepackage{amsmath}
\usepackage{amssymb}
\usepackage{enumerate}
\usepackage{centernot}
\usepackage{multicol,comment}
\usepackage{xspace}
\usepackage{enumitem}
\usepackage{environ}
\usepackage{times}
\usepackage{float}
\usepackage{algorithmicx}
\usepackage{algpseudocode}
\usepackage{bm}
\usepackage{lscape}
\usepackage{pdflscape}
\usepackage{wrapfig}
\usepackage{rotating}
\usepackage{epstopdf}
\usepackage{setspace}
\usepackage{breqn}

\usepackage{epsfig}
\usepackage{xcolor}

\usepackage{algorithm}

\usepackage[hyphens]{url}

\IEEEoverridecommandlockouts

\algnewcommand\algorithmicinput{\textbf{Input:}}
\algnewcommand\INPUT{\item[\algorithmicinput]}
\algnewcommand\algorithmicoutput{\textbf{Output:}}
\algnewcommand\OUTPUT{\item[\algorithmicoutput]}
\algnewcommand\algorithmicproc{\textbf{Procedure}}
\algnewcommand\Proc{\item[\algorithmicproc]}

\usepackage{graphicx,subcaption}

\usepackage[unicode,pdfstartview=FitH]{hyperref}
\makeatletter

\newcommand{\lyxmathsym}[1]{\ifmmode\begingroup\def\b@ld{bold}
	\text{\ifx\math@version\b@ld\bfseries\fi#1}\endgroup\else#1\fi}
\theoremstyle{plain}

{\protect\lemmaname}

{\protect\claimname}

{\protect\corname}

{\protect\defname}

\providecommand{\lemmaname}{Lemma}
\providecommand{\theoremname}{Theorem}
\providecommand{\claimname}{Claim}
\providecommand{\corname}{Corollary}
\providecommand{\defname}{Definition}

\newcommand{\pers}{\texttt{Perspective}\xspace}

\makeatother

\begin{document}
\title{Deceiving Google's \pers API Built for Detecting Toxic Comments}

\author{
	\IEEEauthorblockN{Hossein Hosseini, Sreeram Kannan, Baosen Zhang and Radha Poovendran}\\
	Network Security Lab (NSL), Department of Electrical Engineering, University of Washington, Seattle, WA \\
	Email: \{hosseinh, ksreeram, zhangbao, rp3\}@uw.edu
	\thanks{This work was supported by ONR grants N00014-14-1-0029 and N00014-16-1-2710, ARO grant W911NF-16-1-0485 and NSF grant CNS-1446866.}
}

\maketitle
\thispagestyle{plain}
\pagestyle{plain}
	
\begin{abstract}
Social media platforms provide an environment where people can freely engage in discussions. Unfortunately, they also enable several problems, such as online harassment. Recently, Google and Jigsaw started a project called \pers, which uses machine learning to automatically detect toxic language. A demonstration website has been also launched, which allows anyone to type a phrase in the interface and instantaneously see the toxicity score~\cite{api}. 

In this paper, we propose an attack on the \pers toxic detection system based on the adversarial examples. We show that an adversary can subtly modify a highly toxic phrase in a way that the system assigns significantly lower toxicity score to it. We apply the attack on the sample phrases provided in the \pers website and show that we can consistently reduce the toxicity scores to the level of the non-toxic phrases. The existence of such adversarial examples is very harmful for toxic detection systems and seriously undermines their usability. 
\end{abstract}
\section{Introduction}

Social media platforms provide an environment where people can learn about the trends and news, freely share their opinions and engage in discussions. Unfortunately, the lack of a moderating entity in these platforms has caused several problems, ranging from the wide spread of fake news to online harassment~\cite{duggan2014online}. Due to the growing concern about the impact of online harassment on the people's experience of the Internet, many platforms are taking steps to enhance the safety of the online environments~\cite{Wiki,nytimes2}. 

Some of the platforms employ approaches such as refining the information based on crowdsourcing (upvotes/downvotes), turning off comments or manual moderation to mitigate the effect of the inappropriate contents~\cite{wired}. These approaches however are inefficient and not scalable. As a result, there has been many calls for researchers to develop methods to automatically detect abusive or toxic context in the real time~\cite{wulczyn2016ex}. 

Recent advances in machine learning have transformed many domains such as computer vision~\cite{krizhevsky2012imagenet}, speech recognition~\cite{dahl2012context}, and language processing~\cite{collobert2008unified}. 
Many researchers have explored using machine learning to also tackle the problem of online harassment. Recently, Google and Jigsaw launched a project called \pers~\cite{api}, which uses machine learning to automatically detect online insults, harassment, and abusive speech. The system intends to bring Conversation AI to help with providing a safe environment for online discussions~\cite{jigsaw}.

\pers is an API that enables the developers to use the toxic detector running on Google's servers, to identify harassment and abuse on social media or more efficiently filtering invective from the comments on a news website. Jigsaw has partnered with online communities and publishers, such as Wikipedia~\cite{Wiki} and The New York Times~\cite{nytimes}, to implement this toxicity measurement system. 

Recently, a demonstration website has been launched, which allows anyone to type a phrase in the \pers's interface and instantaneously see how it rates on the ``toxicity'' scale~\cite{api}. The \pers website has also open sourced the experiments, models and research data in order to explore the strengths and weaknesses of using machine learning as a tool for online discussion. 

The implicit assumption of learning models is that they will be deployed in benign settings. However, many works have pointed out their vulnerability in adversarial scenarios~\cite{barreno2006can,barreno2010security,huang2011adversarial}. One type of the vulnerabilities of machine learning algorithms is that an adversary can change the algorithm output by subtly perturbing the input, often unnoticeable by humans. Such inputs are called {\it adversarial examples}~\cite{szegedy2013intriguing}, and have been shown to be effective against different machine learning algorithms even when the adversary has only a black-box access to the target model~\cite{papernot2016practical}.

In this paper, we demonstrate the vulnerability of the recently-released Google's \pers system against the adversarial examples. In the text classification task of the \pers, adversarial examples can be defined as modified texts which contain the same highly abusive language as the original text, yet receive a significantly lower toxicity score from the learning model. Through different experiments, we show that an adversary can deceive the system by misspelling the abusive words or by adding punctuations between the letters. The existence of adversarial examples is very harmful for toxic detector systems and seriously undermines their usability, especially since these systems are likely to be employed in adversarial settings. We conclude the paper by proposing some countermeasures to the proposed attack. 

\begin{table*}[t]
	\centering
	\caption{Demosntration of the Attack on the \pers Toxic Detection System \label{attack}. All phrases in the first column of the table are chosen from the examples provided by the \pers website~\cite{api}.}
	\begin{tabular}{ |p{8.5cm}||p{8.5cm}| } 
		\hline
		{\bf Original Phrase (Toxicity Score)} & {\bf Modified Phrase (Toxicity Score)} \\ 
		\hline
		\hline
		Climate change is happening and it's not changing in our favor. If you think differently you're an {\bf idiot}. (84\%) &  Climate change is happening and it's not changing in our favor. If you think differently you're an {\bf idiiot}. (20\%)  \\ 
		\hline
		They're {\bf stupid}, it's getting warmer, we should enjoy it while it lasts (86\%) &  They're {\bf st.upid}, it's getting warmer, we should enjoy it while it lasts (2\%)  \\ 
		\hline
		They are liberal {\bf idiots} who are {\bf uneducated} (90\%)  &  They are liberal {\bf i.diots} who are {\bf un.educated} (15\%)  \\ 
		\hline
		\hline
		
		{\bf idiots}. backward thinking people. {\bf nationalists}. not accepting facts. susceptible to {\bf lies}. (80\%) &   {\bf idiiots}. backward thinking people. {\bf nationaalists}. not accepting facts. susceptible to {\bf l.ies}. (17\%)  \\ 
		\hline
		They are {\bf stupid} and {\bf ignorant} with no class (91\%)  &   They are {\bf st.upid} and {\bf ig.norant} with no class (11\%)  \\ 
		\hline
		It's {\bf stupid} and wrong (89\%) &  It's {\bf stuipd} and wrong (17\%) \\
		\hline
		\hline
		If they voted for Hilary they are {\bf idiots} (90\%) &  If they voted for Hilary they are {\bf id.iots} (12\%) \\
		\hline
		Anyone who voted for Trump is a {\bf moron} (80\%)  &   Anyone who voted for Trump is a {\bf mo.ron} (13\%)  \\ 
		\hline
		{\bf Screw} you trump supporters (79\%) &   {\bf S c r e w}  you trump supporters (17\%) \\
		\hline
	\end{tabular}
\end{table*}

\section{Background}

\subsection{Brief Description of Google's \pers API}

\pers is an API created by Jigsaw and Google's Counter Abuse Technology team in Conversation-AI. Conversation AI is a collaborative research effort exploring ML as a tool for better discussions online~\cite{conversationai}. The API uses machine learning models to score the toxicity of an input text, where toxic is defined as ``a rude, disrespectful, or unreasonable comment that is likely to make one leave a discussion.''

Google and Jigsaw developed the measurement tool by taking millions of comments from different publishers and then asking panels of ten people to rate the comments on a scale from ``very toxic'' to ``very healthy'' contribution. The resulting judgments provided a large set of training examples for the machine learning model. 

Jigsaw has partnered with online communities and publishers to implement the toxicity measurement system. Wikipedia use it to perform a study of its editorial discussion pages~\cite{Wiki} and The New York Times is planning to use it as a first pass of all its comments, automatically flagging abusive ones for its team of human moderators~\cite{nytimes}. The API outputs the scores in real-time, so that publishers can integrate it into their website to show toxicity ratings to commenters even during the typing~\cite{wired}. 

\subsection{Adversarial Examples for Learning Systems}

Machine learning models are generally designed to yield the best performance on clean data and in benign settings. As a result, they are subject to attacks in adversarial scenarios~\cite{barreno2006can,barreno2010security,huang2011adversarial}. One type of the vulnerabilities of the machine learning algorithms is that an adversary can change the algorithm prediction score by perturbing the input slightly, often unnoticeable by humans. Such inputs are called {\it adversarial examples}~\cite{szegedy2013intriguing}.

Adversarial examples have been applied to models for different tasks, such as images classification~\cite{szegedy2013intriguing,goodfellow2014explaining,papernot2016limitations}, music
content analysis~\cite{kereliuk2015deep} and malware classification~\cite{grosse2016adversarial}. In this work, we generate adversarial examples on a real-world text classifier system. In the context of scoring the toxicity, adversarial examples can be defined as modified phrases that contain the same highly abusive language as the original one, yet receive a significantly lower toxicity score by the model. 

In a similar work~\cite{reddy2016obfuscating}, the authors presented a method for gender obfuscating in social media writing. The proposed method modifies the text such that the algorithm classifies the writer gender as a certain target gender, under limited knowledge of the classifier and while preserving the text's fluency and meaning. 
The modified text is not required to be adversarial, i.e., a human may also classify it as the target gender. In contrast, in the application of toxic text detection, the adversary intends to deceive the classifier, while {\it maintaining the abusive content of the text}.

\begin{table*}[t]
	\centering
	\caption{Demosntration of False Alarm on the \pers Toxic Detection System \label{false}. All phrases in the first column of the table are chosen from the examples provided by the \pers website~\cite{api}}.
	\begin{tabular}{ |p{8.5cm}||p{8.5cm}| } 
		\hline
		{\bf Original Phrase (Toxicity Score)} & {\bf Modified Phrase (Toxicity Score)} \\ 
		\hline
		\hline
		Climate change is happening and it's not changing in our favor. If you think differently you're an idiot (84\%) &  Climate change is happening and it's not changing in our favor. If you think differently you're {\bf not} an idiot (73\%)  \\ 
		\hline
		They're stupid, it's getting warmer, we should enjoy it while it lasts (86\%) &  They're {\bf not} stupid, it's getting warmer, we should enjoy it while it lasts (74\%)  \\ 
		\hline
		They are liberal idiots who are uneducated. (90\%)  &  They are {\bf not} liberal idiots who are uneducated. (83\%)  \\ 
		\hline
		\hline
		idiots. backward thinking people. nationalists. not accepting facts. susceptible to lies. (80\%) &   {\bf not} idiots. {\bf not} backward thinking people. {\bf not} nationalists. accepting facts. {\bf not} susceptible to lies. (74\%)  \\ 
		\hline
		They are stupid and ignorant with no class (91\%)  &   They are {\bf not} stupid and ignorant with no class (84\%)  \\ 
		\hline
		It's stupid and wrong (89\%) &  It's {\bf not} stupid and wrong (83\%) \\
		\hline
		\hline
		If they voted for Hilary they are idiots (90\%) &  If they voted for Hilary they are {\bf not} idiots (81\%) \\
		\hline
		Anyone who voted for Trump is a moron (80\%)  &   Anyone who voted for Trump is {\bf not} a moron (65\%)  \\ 
		\hline
		Screw you trump supporters (79\%) &   {\bf Will not} screw you trump supporters (68\%) \\
		\hline
	\end{tabular}
\end{table*}

\section{The Proposed Attacks}\label{sec:attack}

Recently, a website has been launched for \pers demonstration, which allows anyone to type a phrase in the interface and instantaneously receive its toxicity score~\cite{api}. The website provides samples phrases for three categories of topics ``that are often difficult to discuss online''. The categories are 1) Climate Change, 2) Brexit and 3) US Election. 

In this section, we demonstrate an attack on the \pers toxic detection system, based on the adversarial examples. In particular, we show that an adversary can subtly modify a toxic phrase such that the model will output a very low toxicity score for the modified phrase. The attack setting is as follows. The adversary possesses a phrase with a toxic content and tries different perturbations on the words, until she succeeds with significantly reducing the confidence of the model that the phrase is toxic. Note that the adversary does not have access to the model or training data, and can only query the model and get the toxicity score.

Table~\ref{attack} demonstrates the attack on sample phrases provided by the \pers website. The first column represents the original phrases along with the toxicity scores and the second column provides the adversarially modified phrases and their corresponding toxicity scores. 
\footnote{The experiments are done on the interface of the \pers website on Feb. 24, 2017.}
For better demonstration of the attack, we chose phrases with different toxic words and also introduced different types of errors, rather than searching for the best error type that would potentially yield lower toxicity score. The boldface words are the toxic words that the adversary has modified. The modifications are adding a dot between two letters, adding spaces between all letters or misspelling the word (repeating one letter twice or swapping two letters). As can be seen, we can consistently reduce the toxicity score to the level of the benign phrases by subtly modifying the toxic words. 

Moreover, we observed that the adversarial perturbations {\it transfer} among different phrases, i.e., if a certain modification to a word reduces the toxicity score of a phrase, the same modification to the word is likely to reduce the toxicity score also for another phrase. Using this property, an adversary can form a dictionary of the adversarial perturbations for every word and significantly simplify the attack process. 

Through the experiments, we made the following observations:
\begin{itemize}
	\item Susceptibility to false alarm: we observed that the \pers system also wrongly assigns high toxicity scores to the apparently benign phrases. Table~\ref{false} demonstrates the false alarm on the same sample phrases of Table~\ref{attack}. The first column represents the original phrases along with the toxicity scores and the second column provides the negated phrases and the corresponding toxicity scores. The boldface words are added to toxic phrases. As can be seen, the system consistently fails to capture the inherent semantic of the modified phrases and wrongly assigns high toxicity scores to them.
	\item Robustness to random misspellings: we observed that the system assigns 34\% toxicity score to most of the misspelled and random words. Also, it is somewhat robust to phrases that contain randomly modified toxic words.
	\item Vulnerability to poisoning attack: The \pers interface allows users to provide a feedback on the toxicity score of  phrases, suggesting that the learning algorithm updates itself using the new data. This can expose the system to poisoning attacks, where an adversary modifies the training data (in this case, the labels) so that the model assigns low toxicity scores to certain phrases.
\end{itemize}

\section{Open Problems in Defense Methods}

The developers of \pers have mentioned that the system is in the early days of research and development, and that the experiments, models, and research data are published to explore the strengths and weaknesses of using machine learning as a tool for online discussion. 


In section~\ref{sec:attack}, we showed the vulnerability of the \pers system against the adversarial examples. Scoring the semantic toxicity of a phrase is clearly a very challenging task. In this following, we briefly review some of the possible approaches for improving the robustness of the toxic detection systems:

\begin{itemize}
	\item Adversarial Training: In this approach, during the training phase, we generate the adversarial examples and train the model to assign the original label to them~\cite{goodfellow2014explaining}. In the context of toxic detection systems, we need to include different modified versions of the toxic words into the training data. While this approach may improve the robustness of the system against the adversarial examples, it does not seem practical to train the model on all variants of every word.
	
	\item Spell checking: Many of the adversarial examples can be detected by first applying a spell checking filter before the toxic detection system. This approach may however increase the false alarm. 
	
	\item Blocking suspicious users for a period of time: The adversary needs to try different error patterns to finally evade the toxic detection system. Once a user fails to pass the threshold for a number of times, the system can block her for a while. This approach can force the users to less often use toxic language.
	
\end{itemize}

\section{Conclusion}

In this paper, we presented an attack on the recently-released Google's \pers API built for detecting toxic comments. We showed that the system can be deceived by slightly perturbing the abusive phrases to receive very low toxicity scores, while preserving the intended meaning. We also showed that the system has high false alarm rate in scoring high toxicity to benign phrases. We provided detailed examples for the studied cases. Our future work includes development of countermeasures against such attacks.


\noindent{\bf Disclaimer:} The phrases used in Tables~\ref{attack} and~\ref{false} are chosen from the examples provided in the \pers website~\cite{api} for the purpose of demonstrating the results and do not represent the view or opinions of the authors or sponsoring agencies. 

\bibliographystyle{ieeetr}
\bibliography{Main}

\begin{thebibliography}{10}

\bibitem{api}
``{https://www.perspectiveapi.com/},''

\bibitem{duggan2014online}
M.~Duggan, {\em Online harassment}.
\newblock Pew Research Center, 2014.

\bibitem{Wiki}
``{https://meta.wikimedia.org/wiki/Research:Detox},''

\bibitem{nytimes2}
``{https://www.nytimes.com/interactive/2016/09/20/insider/approve-or-reject-moderation-quiz.html},''

\bibitem{wired}
``{https://www.wired.com/2017/02/googles-troll-fighting-ai-now-belongs-world/},''

\bibitem{wulczyn2016ex}
E.~Wulczyn, N.~Thain, and L.~Dixon, ``Ex machina: Personal attacks seen at
  scale,'' {\em arXiv preprint arXiv:1610.08914}, 2016.

\bibitem{krizhevsky2012imagenet}
A.~Krizhevsky, I.~Sutskever, and G.~E. Hinton, ``Imagenet classification with
  deep convolutional neural networks,'' in {\em Advances in neural information
  processing systems}, pp.~1097--1105, 2012.

\bibitem{dahl2012context}
G.~E. Dahl, D.~Yu, L.~Deng, and A.~Acero, ``Context-dependent pre-trained deep
  neural networks for large-vocabulary speech recognition,'' {\em IEEE
  Transactions on Audio, Speech, and Language Processing}, vol.~20, no.~1,
  pp.~30--42, 2012.

\bibitem{collobert2008unified}
R.~Collobert and J.~Weston, ``A unified architecture for natural language
  processing: Deep neural networks with multitask learning,'' in {\em
  Proceedings of the 25th international conference on Machine learning},
  pp.~160--167, ACM, 2008.

\bibitem{jigsaw}
``{https://jigsaw.google.com/},''

\bibitem{nytimes}
``{http://www.nytco.com/the-times-is-partnering-with-jigsaw-to-expand-comment-capabilities/},''

\bibitem{barreno2006can}
M.~Barreno, B.~Nelson, R.~Sears, A.~D. Joseph, and J.~D. Tygar, ``Can machine
  learning be secure?,'' in {\em Proceedings of the 2006 ACM Symposium on
  Information, computer and communications security}, pp.~16--25, ACM, 2006.

\bibitem{barreno2010security}
M.~Barreno, B.~Nelson, A.~D. Joseph, and J.~Tygar, ``The security of machine
  learning,'' {\em Machine Learning}, vol.~81, no.~2, pp.~121--148, 2010.

\bibitem{huang2011adversarial}
L.~Huang, A.~D. Joseph, B.~Nelson, B.~I. Rubinstein, and J.~Tygar,
  ``Adversarial machine learning,'' in {\em Proceedings of the 4th ACM workshop
  on Security and artificial intelligence}, pp.~43--58, ACM, 2011.

\bibitem{szegedy2013intriguing}
C.~Szegedy, W.~Zaremba, I.~Sutskever, J.~Bruna, D.~Erhan, I.~Goodfellow, and
  R.~Fergus, ``Intriguing properties of neural networks,'' {\em arXiv preprint
  arXiv:1312.6199}, 2013.

\bibitem{papernot2016practical}
N.~Papernot, P.~McDaniel, I.~Goodfellow, S.~Jha, Z.~B. Celik, and A.~Swami,
  ``Practical black-box attacks against deep learning systems using adversarial
  examples,'' {\em arXiv preprint arXiv:1602.02697}, 2016.

\bibitem{conversationai}
``{https://conversationai.github.io/},''

\bibitem{goodfellow2014explaining}
I.~J. Goodfellow, J.~Shlens, and C.~Szegedy, ``Explaining and harnessing
  adversarial examples,'' {\em arXiv preprint arXiv:1412.6572}, 2014.

\bibitem{papernot2016limitations}
N.~Papernot, P.~McDaniel, S.~Jha, M.~Fredrikson, Z.~B. Celik, and A.~Swami,
  ``The limitations of deep learning in adversarial settings,'' in {\em 2016
  IEEE European Symposium on Security and Privacy (EuroS\&P)}, pp.~372--387,
  IEEE, 2016.

\bibitem{kereliuk2015deep}
C.~Kereliuk, B.~L. Sturm, and J.~Larsen, ``Deep learning and music
  adversaries,'' {\em IEEE Transactions on Multimedia}, vol.~17, no.~11,
  pp.~2059--2071, 2015.

\bibitem{grosse2016adversarial}
K.~Grosse, N.~Papernot, P.~Manoharan, M.~Backes, and P.~McDaniel, ``Adversarial
  perturbations against deep neural networks for malware classification,'' {\em
  arXiv preprint arXiv:1606.04435}, 2016.

\bibitem{reddy2016obfuscating}
S.~Reddy, M.~Wellesley, K.~Knight, and C.~Marina~del Rey, ``Obfuscating gender
  in social media writing,'' {\em NLP+ CSS 2016}, p.~17, 2016.

\end{thebibliography}

\end{document}